\documentclass[letterpaper, 10 pt, conference]{ieeeconf}
\IEEEoverridecommandlockouts
\usepackage{algorithm}
\usepackage{algpseudocode}
\usepackage{graphicx}
\usepackage{svg}
\usepackage{subcaption}
\usepackage[font=small]{caption}
\usepackage{siunitx}

\usepackage[
    backend=bibtex,
    style=ieee,
    mincitenames=1,
    maxcitenames=2,
    maxbibnames=6,
    isbn=false,
    url=false,
    doi=false,
    natbib=true]{biblatex}
\usepackage{amsmath,amssymb,amsfonts}
\usepackage{textcomp}
\usepackage[table]{xcolor}
\usepackage{booktabs}
\usepackage{tabularx}
\usepackage{multicol}
\usepackage{svg}
\usepackage[bookmarks=true]{hyperref}
\usepackage[capitalise,nameinlink]{cleveref} 
\usepackage{todonotes}
\def\BibTeX{{\rm B\kern-.05em{\sc i\kern-.025em b}\kern-.08em
    T\kern-.1667em\lower.7ex\hbox{E}\kern-.125emX}}

\begin{document}

\title{\LARGE \bf\textit{ContactKernels}: Adaptive Proposals using Geometry-Conditioned Wrenches for Contact-Rich Manipulation}

\author{
Aditya Kamireddypalli$^1$, Mat{\'i}as Mattamala$^{1}$, Jo\~{a}o Moura$^1$, Sethu Vijayakumar$^1$, \\Steve Tonneau$^1$, Subramanian Ramamoorthy$^1$
%
\thanks{$^1$School of Informatics, University of Edinburgh, Scotland. $\texttt{a.kamireddypalli@sms.ed.ac.uk}$}
\thanks{$^2$Department of Mechanical and Mechatronics Engineering, University of Waterloo, Canada.}
\thanks{\noindent A. Kamireddypalli is supported in part by funding in the form of an unrestricted gift from Microsoft Research. S. Ramamoorthy is supported by a UKRI Turing AI World Leading Researcher Fellowship on AI for Person-Centred and Teachable Autonomy (grant EP/Z534833/1).}
\thanks{\noindent For the purpose of open access, the authors have applied a Creative Commons Attribution (CC BY) license to any Author Accepted Manuscript version arising from this submission.}
}

\maketitle


\begin{abstract}
Contact-rich insertion requires precise geometric alignment despite uncertainty in perception, geometry, and control.
When this uncertainty causes unintended contact, the resulting contact provides local information that can guide corrective motion.
Sampling-based Model Predictive Control (SMPC) is well suited to this setting because it optimizes through forward simulation without requiring gradients through contact dynamics.
However, standard SMPC methods typically use generic proposal distributions that do not exploit observed contact to determine where to search next.
We introduce \textit{ContactKernels}, a contact-adaptive proposal kernel that uses geometry-conditioned wrench measurements to estimate contact location and local contact geometry.
The estimated contact state directly adapts the proposal distribution: contact location biases the search toward corrective motions, while the contact normal shapes the sampling covariance.
We evaluate \textit{ContactKernels} with Predictive Sampling (PS) and Model Predictive Path Integral (MPPI) control across multiple insertion geometries in simulation and on a real robot.
In simulation, \textit{ContactKernels} improves mean insertion success by over \SI{50}{\percent} points across both controllers.
On the real robot, it increases insertion success from \SI{53.3}{\percent} to \SI{90.0}{\percent} and reduces mean time to insertion by \SI{33}{\percent}.
These results show that contact information can improve sampling-based control not only as feedback, but also by directly determining where the controller searches.
\end{abstract}

\section{Introduction}

Insertion is a fundamental manipulation task in which small uncertainties in geometry, perception, or control can lead to unexpected contact~\cite{jiang_review_2022}.
Successful insertion requires precise alignment between the manipulated object and its environment.
Tight geometric tolerances make this alignment difficult in the presence of uncertainty in perception, local geometry, and control execution~\cite{lozano1984automatic}.
This uncertainty can cause unintended contact and prevent insertion.

Contact, however, is not only a constraint.
It also provides information about the local geometric mismatch.
In particular, sensed contact can provide information about where the manipulated object contacts its environment.
Humans use this information to guide corrective motion.
When inserting a key, for example, we often “wiggle” the key after contact to search for a initial point of a potential feasible insertion path.
A robot can similarly use sensed contact location to determine where to explore next.

This has motivated the development of controllers that can use contact information to adapt its motion online~\cite{lee_making_2019, kim_active_2022}.
Learning-based methods can react to contact, but acquire the contact-to-action mapping from prior experience~\cite{luo_serl_2024, luo_precise_2025}.
Trajectory optimization can instead reason explicitly about contact at test time, but it requires an optimization formulation that handles nonlinear and nonsmooth contact dynamics~\cite{Shirai2023RobustOptimization}.
We seek an alternative that can use contact to adapt its search online without requiring a learned contact-to-action mapping or gradients through contact dynamics.

Sampling-based Model Predictive Control (SMPC) provides this capability.
SMPC searches for control trajectories through forward simulation and does not require gradients through contact dynamics.
However, standard SMPC methods typically sample controls from generic proposal distributions.
After contact, these proposals might continue to sample motions that are inconsistent with the observed contact state, because contact information has not been used in the sampling process.
\begin{figure}
    \centering
    \includegraphics[width=1.0\linewidth]{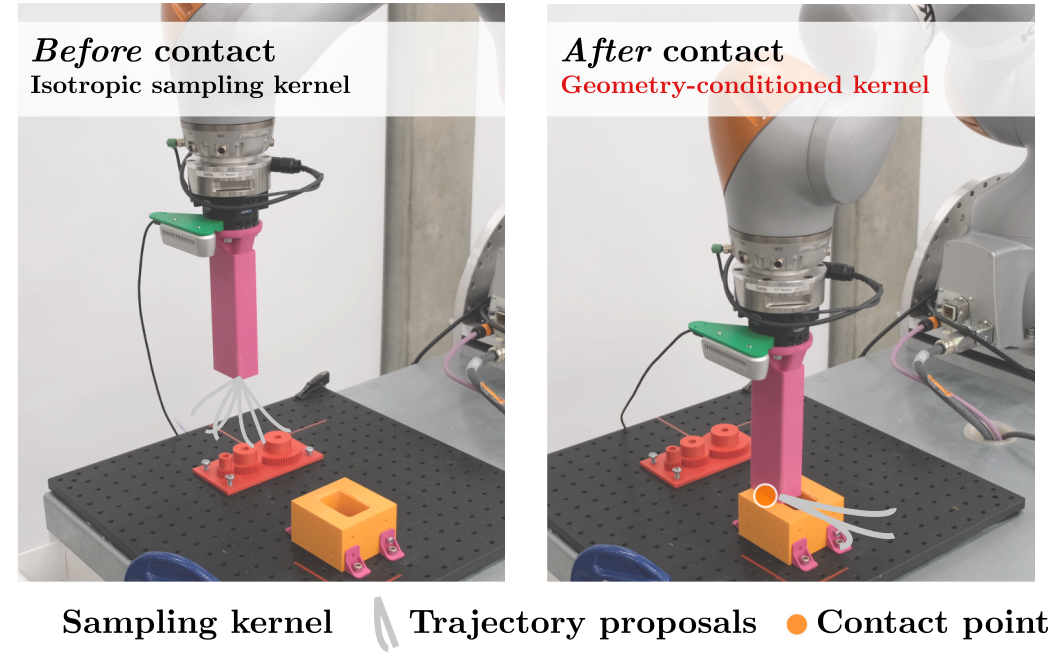}
    \caption{\textbf{ContactKernels}: In this work we propose a Sampling-based MPC approach to the peg insertion task. To efficiently perform insertion, we utilise geometry-conditioned wrenches to steer sampling via an adaptive proposal kernel. This is shown to outperform baseline isotropic Gaussian Kernel.}
    \label{fig:hero-figure}
\end{figure}

We address this gap with \emph{ContactKernels}, a contact-adaptive proposal kernel for sampling-based MPC.
\emph{ContactKernels} uses geometry-conditioned wrench measurements to estimate the contact location and local contact geometry.
The estimated contact location biases the proposal toward corrective motions, while the contact normal shapes the sampling covariance.
Contact confidence scales the strength of this adaptation, directly determining where the controller should search after unintended contact.

The contributions of this work are threefold:
\begin{enumerate}
    \item We formulate contact-rich insertion using sampling-based Model Predictive Control (SMPC) and implement the controller with both Predictive Sampling (PS) and Model Predictive Path Integral (MPPI) control.

    \item We introduce \textit{ContactKernels}, an adaptive proposal kernel that uses estimated contact location and local contact geometry to guide where the controller samples. In simulation, \textit{ContactKernels} increases mean insertion success from $15.8\%$ to $66.7\%$ with PS and from $13.5\%$ to $65.6\%$ with MPPI.

    \item We validate the approach across multiple insertion geometries in simulation and on a real robot. On the robot, MPPI with \textit{ContactKernels} increases insertion success from $53.3\%$ to $90.0\%$ and reduces the mean time to insertion by $33\%$.
\end{enumerate}

\section{Related Work}
This section discusses predominant peg-insertion strategies, application of SMPC for contact-rich manipulation tasks and methods that have tracked and closed the loop on contact state.

\subsection{Contact-rich Insertion Strategies}

Contact-rich insertion requires the robot to adapt when uncertainty causes unintended contact~\cite{jiang_review_2022}.
Existing methods use model-based or learning-based strategies.

Model-based methods explicitly represent contact between the robot and its environment.
Classical methods use geometric and contact models to define corrective or compliant motions based on contact mode detection~\cite{whitney_quasi-static_1982, shirinzadeh_hybrid_2011, usubamatov_analyses_2011}.

Alternately, Trajectory Optimization (TO) directly estimates parameters of a trajectory subject to dynamics and kinematics constraints.
At suitable high frequencies, TO utilised in a receding horizon MPC loop generates trajectories that potentially adapt to unintended contact.
While successfully applied to contact-rich pushing tasks~\cite{Aydinoglu2023ConsensusMPC, venkatesh_approximating_2025}, and tool manipulation~\cite{Shirai2023RobustOptimization, Shirai2023TactileManipulation} TO has seen limited application to the problem of contact-rich insertion.
Optimization-based methods must handle nonlinear and nonsmooth contact dynamics, exacerbated by changing contact modes for the insertion model.
Recent contact-implicit methods include smooth approximations to contact dynamics within an optimal control problem~\cite{dietz_robust_2026}. 
These methods can reason about contact at deployment time.
Contact-rich trajectory optimization typically requires additional modeling machinery—such as complementarity relaxation, smoothing, or specialized differentiable simulation—to obtain informative gradients across changing contact modes.
ContactKernels instead avoids this requirement by performing zero-order optimization directly over simulated interactions and using realized contact to adapt the sampling distribution.

To circumvent the limitations of model specification, learning-based methods learn contact-dependent behaviours from data.
Learning-from-Demonstration (LfD) learns insertion strategies from expert demonstrations~\cite{ijspeert_movement_2002, hovland_skill_1996}.
Reinforcement learning instead learns through interaction in simulation or the real world~\cite{levine_learning_2015, fan_learning_2019, noseworthy_forge_2025, tang_industreal_2023}.
Demonstrations can also bootstrap reinforcement learning and improve sample efficiency~\cite{davchev_residual_2022, levine_learning_2015, schoettler_deep_2020, luo_serl_2024}.
Recent methods use human corrective feedback to further improve exploration~\cite{luo_precise_2025}.
These methods can react to contact online.
However, the contact-to-action mapping is learned from prior experience.
Generalisation to unseen contact conditions therefore depends on the learned policy.
When this fails, adaptation requires further interaction, expert demonstrations, or policy fine-tuning.

Model-based and learning-based approaches thus provide complementary mechanisms for contact adaptation. Model-based methods explicitly reason about contact, but require an appropriate interaction model and control or optimization formulation. Learning-based methods can represent complex contact-dependent behaviours, but acquire the adaptation mechanism from prior experience.

Our approach instead estimates contact location and local contact geometry during execution.
We use this information within a zero-order optimization framework, where the estimated contact state directly changes where the controller searches for subsequent motions.
This enables test-time contact adaptation without additional policy training or differentiation through contact dynamics.

\subsection{Sampling-based MPC for Contact-rich Manipulation}
Sampling-based model predictive control (SMPC) performs online trajectory optimization by evaluating sampled control sequences through forward simulation~\cite{theodorou_generalized_2010, williams_model_2017}.
Rather than relying on gradients of the system dynamics, these methods optimize control by comparing the predicted performance of candidate trajectories~\cite{jordana_introduction_2025}.
This makes them particularly attractive for contact-rich manipulation, where dynamics is nonlinear, hybrid, and difficult to differentiate due to intermittent contact, friction, and mode transitions.
When paired with modern parallelised simulation tools~\cite{todorov_mujoco_2012}, SMPC becomes a powerful paradigm for contact-rich robot control~\cite{Howell2022PredictiveMuJoCo, pezzato_sampling-based_2025}.

Recent work has demonstrated this effectively for pushing and dextrous manipulation by including GPU-parallel physics rollouts inside the MPC loop~\cite{pezzato_sampling-based_2025}.
However, the literature for applying SMPC to the peg insertion task remains limited.
MPQ($\lambda$)~\cite{bhardwaj_blending_2021} and CaPTURe~\cite{marques_particle-based_2026} tackle transition model bias that arises from chosen approximation strategies, and uncertainty associated with changing contact modes in peg insertion setups respectively.
Both incorporate partially learned dynamics as an offline step, and do not evaluate on a real robot.
This leaves real-world peg insertion using parallel sampling simulation-based MPC largely unexplored.

However, applying SMPC to insertion does not by itself exploit the contact information acquired during execution.
Standard SMPC methods typically sample controls from generic proposal distributions~\cite{hess_sampling-based_2025, pezzato_sampling-based_2025}.
These distributions do not use the observed contact state to determine where to explore next.
This can lead to inefficient exploration after unintended contact.
ContactKernels addresses this gap by conditioning the proposal distribution on the estimated contact location and local geometry.
Contact therefore directly guides where the controller searches for subsequent motions.

\subsection{Closing the Loop on Contact State}

%
%

Integrating contact state information can improve adaptation during contact-rich insertion~\cite{lee_making_2019, pares-morlans_demystifying_2026, kim_active_2022}.
Learning-based methods can incorporate tactile observations directly into a feedback policy~\cite{lee_making_2019}.
These methods can learn complex contact-dependent behaviours from high-dimensional sensory inputs.
However, the contact-to-action mapping is acquired during training.
The resulting adaptation therefore depends on the contact behaviours represented by the learned policy.

An alternative is to explicitly estimate a task-relevant contact state before selecting an action.
Kim and Rodriguez~\cite{kim_active_2022} estimate the extrinsic contact line between a grasped object and its environment.
An active tactile controller regulates the contact mode to improve this estimate.
The estimated contact line is then provided to an object-agnostic insertion policy trained in simulation.
This structured representation enables the policy to generalize across different object and hole geometries.

These approaches demonstrate the benefit of closing the loop on contact state.
However, the estimated contact state is still mapped to actions through a learned policy.
We instead use contact state to adapt the search process of an online optimizer.
ContactKernels estimates contact location and local contact geometry from measured wrenches.
This information directly shapes the proposal distribution of the SMPC controller.
Contact therefore determines where the controller searches for subsequent motions, without requiring the contact-to-action mapping itself to be learned.

\section{Background}

We start with providing a brief background on the use of Sampling-based MPC for contact-rich manipulation

\subsection{Sampling-based MPC}

Model Predictive Control (MPC) iteratively calculates future controls using a model of the system.
Let $\mathbf{x}_t$ denote the state and $\mathbf{u}_t$ the control at time $t$, with dynamics being specified by $\mathbf{x}_{t+1}=f(\mathbf{x}_t,\mathbf{u}_t)$.
From the current state $\mathbf{x}_0$, MPC plans a sequence of controls over a horizon $H$ given by $\mathbf{U}=(\mathbf{u}_0,\mathbf{u}_1,\ldots,\mathbf{u}_{H-1})$.
Simulating this sequence through the dynamics model produces a predicted state trajectory, called a \emph{rollout}:
$\mathbf{X}=(\mathbf{x}_0,\mathbf{x}_1,\ldots,\mathbf{x}_H)$.

Each rollout is evaluated using a trajectory cost:
\begin{equation}
\label{eq:running_cost}
J(\mathbf{U})=\sum_{t=0}^{H-1}\ell(\mathbf{x}_t,\mathbf{u}_t)+\ell_f(\mathbf{x}_H),
\end{equation}

where $\ell$ is the running cost and $\ell_f$ is the terminal cost. MPC seeks the control sequence

\begin{equation}
\mathbf{U}^*=\arg\min_{\mathbf{U}} J(\mathbf{U}).
\end{equation}

Only the first control $\mathbf{u}_0^*$ is executed.
The new state is then observed, and the planning problem is solved again.

In sampling-based MPC, the optimizer generates $N$ candidate control sequences,
$\mathbf{U}^{(1)},\mathbf{U}^{(2)},\ldots,\mathbf{U}^{(N)}$, and evaluates each candidate by rolling it out through the dynamics model. The candidates are drawn from a \emph{proposal distribution}, or specified by a \emph{proposal kernel},
\begin{equation}
\mathbf{U}^{(i)} \sim \pi(\mathbf{U}\mid \bar{\mathbf{U}},\mathbf{x}_0).
\end{equation}

Here, $\bar{\mathbf{U}}$ is the nominal control sequence, and $\pi$ determines how candidate control sequences are sampled.
The term \emph{kernel} emphasizes that the proposal may depend on the current nominal sequence.
A common choice is a Gaussian proposal centred on the nominal sequence:
\begin{equation}
\pi(\mathbf{U}\mid \bar{\mathbf{U}})=\mathcal{N}(\bar{\mathbf{U}},\Sigma).
\end{equation}

Equivalently,
\begin{equation}
\mathbf{U}^{(i)}=\bar{\mathbf{U}}+\epsilon^{(i)},
\qquad
\epsilon^{(i)}\sim\mathcal{N}(0,\Sigma).
\end{equation}
The covariance $\Sigma$ controls the scale and correlation of exploration. Larger covariances produce more diverse control sequences, while smaller covariances concentrate samples near the nominal sequence.

Random shooting, or Predictive Sampling (PS)~\cite{Howell2022PredictiveMuJoCo}, selects the lowest-cost candidate directly. The Cross-Entropy Method (CEM)~\cite{jordana_introduction_2025} adapts the mean and covariance of the proposal using an elite subset of low-cost samples. Model Predictive Path Integral control (MPPI)~\cite{williams_model_2017} updates the nominal sequence using a cost-weighted combination of the sampled perturbations.
\section{Method}
\begin{figure}[t]
    \centering
    \includegraphics[width=1\linewidth]{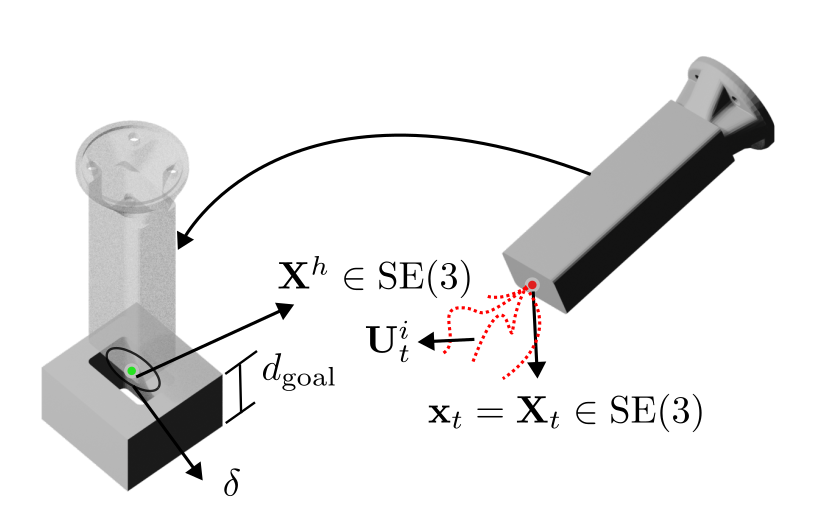}
    \caption{\textbf{Notation Schematic}}
    \label{fig:method}
\end{figure}
\begin{figure}
    \centering
    \vspace{2mm}
    \includegraphics[width=0.9\columnwidth]{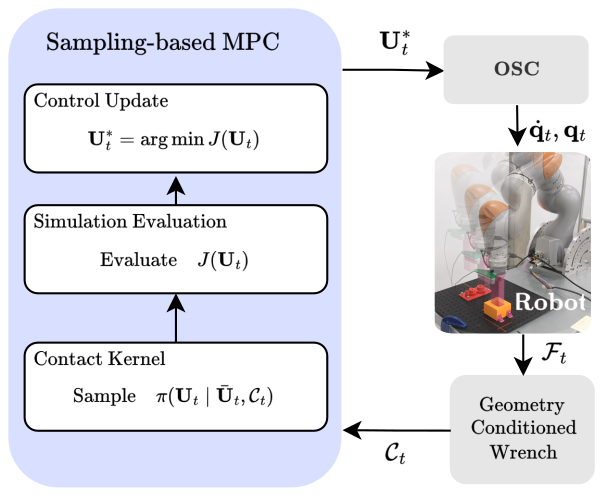}
    \caption{\textbf{System overview}: \emph{ContactKernels} is a Sampling-based MPC framework closing the loop on contact location for contact-rich manipulation. Demonstrated for the peg-insertion task, our framework incorporates contact information via a proposal kernel, before sampling new trajectories. This enables adaptation in fine contact-rich manipulation tasks.}
    \label{fig:system-diagram}
\end{figure}

\subsection{Problem Formulation}

We consider a contact-rich peg-insertion task in which the robot must align a rigidly grasped peg with a fixed hole and subsequently insert it.
We assume a fixtured grasp.
The system state is represented by the peg pose
\begin{equation}
\mathbf{x}_t = \in \mathrm{SE}(3),
\qquad
\mathbf{x}_t = (\mathbf{R}_t,\mathbf{p}_t),
\end{equation}
where $\mathbf{R}_t\in \mathrm{SO}(3)$ and $\mathbf{p}_t\in\mathbb{R}^3$ denote the peg orientation and position.
The robot is tasked to move the peg tip into the hole.
The target hole pose is fixed and denoted by $\mathbf{X}^{h}=\{\mathbf{R}^{h},\mathbf{p}^{h}) \mid \mathbf{R}^h\in \mathrm{SO}(3), \mathbf{p}^h\in\mathbb{R}^3 \}$.
$\mathbf{X}^h$ specifies the pose of the entry of the hole in world frame.

The control input is an end-effector twist,
\begin{equation}
\label{eq:control}
\mathbf{u}_t=
\begin{bmatrix}
\mathbf{v}_t\\
\boldsymbol{\omega}_t
\end{bmatrix}
\in\mathbb{R}^{6},
\end{equation}
where $\mathbf{v}_t$ and $\boldsymbol{\omega}_t$ are the commanded linear and angular velocities.
Over a planning horizon $H$, the control sequence $\mathbf{U}_t$
is represented using $M$ uniformly spaced control knot points, where each
knot is an end-effector twist as defined in~\cref{eq:control}, and cubicly interpolated.

Sampling-based MPC generates $K$ candidate control sequences
$\mathbf{U}_t^{(i)}$ by sampling these control knots from the proposal
distribution.
Each sampled sequence is propagated using through contact dynamics via simulation,
\begin{equation}
    \mathbf{x}_{k+1} = f(\mathbf{x}_k,\mathbf{u}_k),
\end{equation}
and evaluated using the finite-horizon objective
$J(\mathbf{U}_t^{(i)};\mathbf{x}_t)$.
Predictive Sampling selects the minimum-cost sampled sequence,
\begin{equation}
    \mathbf{U}^*_{t}
    =
    \arg\min_{i \in \{1,\ldots,K\}}
    J\!\left(\mathbf{U}_t^{(i)};\mathbf{x}_t\right).
    \label{eq:ps-selection}
\end{equation}
Only the first control of the selected sequence is executed before replanning.
The end-effector twist commands are then mapped to commanded robot joint states via an Operational Space Control (OSC)~\cite{Khatib1987AFormulation}.
%
\cref{fig:system-diagram} describes our overall pipeline.

\subsection{Peg-Insertion Cost}
\label{sec:cost}

The task objective encourages translational alignment, rotational alignment and insertion progress.
We first express the peg-to-hole displacement in coordinates aligned with the current peg frame, given by $\tilde{\mathbf{p}}_t = \mathbf{R}_t^\top (\mathbf{p}^{h}-\mathbf{p}_t)$.
Note that $\tilde{\mathbf{p}}_t$ denotes the the distance between peg tip and the hole entry point expressed in frame coinciding with the peg tip.

The translational alignment term penalizes displacement between peg tip and the hole entry.
\begin{equation}
J_{\text{tr}}(\mathbf{x}_t)
=
\left\|
\tilde{\mathbf{p}}_t
\right\|_2.
\end{equation}
Rotational misalignment is measured using the geodesic distance on $\mathrm{SO}(3)$~\cite{Lynch2017ModernControl},
\begin{equation}
J_{\text{rot}}(\mathbf{x}_t)
=
d_R(\mathbf{R}_t,\mathbf{R}^{h}),
\end{equation}
Once the peg is aligned with the hole entry, we define an insertion progress term that is gated. 
To encourage progress along the insertion axis, we define
\begin{equation}
J_{\text{insert}}(\mathbf{x}_t)
=
|\tilde p_{t,z}-d_{\text{goal}}|,
\end{equation}
where $d_{\text{goal}}$ is the desired displacement along the peg $z$-axis.

The running cost is
\begin{equation}
\begin{aligned}
\ell(\mathbf{x}_k,\mathbf{u}_k)
={}&
w_{\text{tr}} J_{\text{tr}}(\mathbf{x}_k)
+ w_{\text{rot}} J_{\text{rot}}(\mathbf{x}_k) \\
&+ g_k w_{\text{insert}} J_{\text{insert}}(\mathbf{x}_k).
\end{aligned}
\end{equation}
where insertion is activated only when translational alignment is under a threshold $\delta$, and each cost term is associated with respective weights $w_{\text{tr}}, w_{\text{rot}}, w_{\text{insert}}$.
\begin{equation}
g_k
=
\mathbb{I}
\left[
J_{tr}(\mathbf{x}_k)\leq\delta
\right].
\end{equation}
The total trajectory cost is given as per \cref{eq:running_cost}

\subsection{Contact Representation}
\begin{figure}[t]
    \centering
    \vspace{3mm}
    \includegraphics[width=1.0\linewidth]{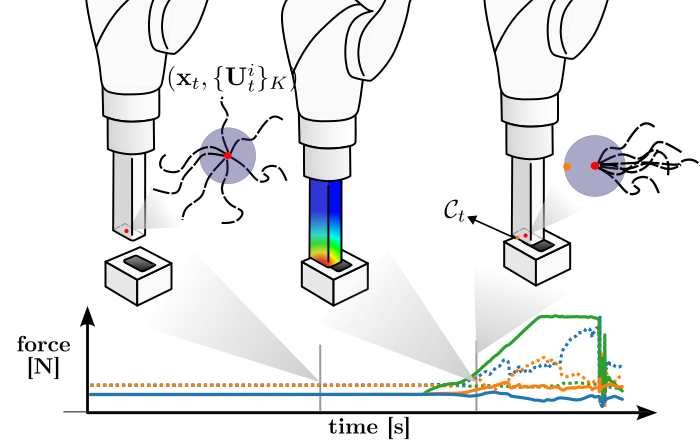}
    \caption{\textbf{Geometry-Conditioned Wrench}: F/T wrenches when conditioned with mesh geometry enables contact location estimation. This is used to adaptively generate trajectory samples upon contact.} 
    \label{fig:contact_location}
\end{figure}

To close the loop on contact location information we define Geometry-conditioned wrenches, where F/T data is used to estimate contact locations.
At each control step, the contact estimator provides $P$ candidate contact points, associated surface normals, and normalized weights,
\begin{equation}
\left\{
\mathbf{p}_i^c,\mathbf{n}_i,w_i
\right\}_{i=1}^{P},
\qquad
\sum_{i=1}^{P} w_i=1.
\end{equation}
The robot senses a wrench $\mathcal{F}_t = \{\mathbf{f}_t, \boldsymbol{\tau}_t\}$ at time $t$.
$P$ points/normals are sampled on the mesh surface of the peg.
The weights of each point is evaluated using the moment-arm model as follows
\begin{align}
   \rho_i &= ||\boldsymbol{\tau}_t - (r_i \times \mathbf{f}_t)||_2 \\
   w_i &= \frac{\exp{\rho_i}}{\sum_P \exp{\rho_i}}
\end{align}
Where $r_i$ is the moment arm between the sampled point and F/T sensor origin, as shown in \cref{fig:contact_location}.
The contact position is estimated as the weighted mean $\mathbf{p}^c_t = \sum_{P}w_i\mathbf{p}^{c}_i$.

To obtain a dominant contact direction without requiring consistent normal orientation, we form the weighted scatter matrix
\begin{equation}
\mathbf{M}_t
=
\sum_{i=1}^{N_c}
w_i\mathbf{n}_i\mathbf{n}_i^\top,
\end{equation}
and define the dominant normal as its principal eigenvector, $\mathbf{n}_t = \operatorname{eig}_{\max}(\mathbf{M}_t)$.
We use the concentration of the particle weights as a confidence measure.
A standard way of doing this is to use the Effective Sample Size metric~\cite{doucet_tutorial_nodate} $N_{\mathrm{ess}} = \frac{1}{\sum_i w_i^2}$,
with the normalized contact confidence specified as,
\begin{equation}
c_t
=
\frac{P-N_{\mathrm{ess}}}{P-1}.
\end{equation}

Finally, we define a \emph{search direction} from the estimated contact location toward the peg tip,
\begin{equation}
\mathbf{s}_t
=
\frac{
\mathbf{p}_t-\mathbf{p}_t^c
}{
\|\mathbf{p}_t-\mathbf{p}_t^c\|_2
}.
\end{equation}
giving us the following contact representation tuple,
\begin{equation}
\mathcal{C}_t
=
(\mathbf{n}_t,c_t,\mathbf{s}_t).
\end{equation}

\subsection{Adaptive Sampling Kernel}
\label{sec:adaptive_proposal}

The contact representation introduced previously is used to adapt the sampling distribution around the nominal sequence $\bar{\mathbf{U}}_t=\mathbf{U}_{t-1}^*$.
The search direction $\mathbf{s}_t$ is used to compute a mean shift to the otherwise baseline isotropic Gaussian kernel.
The contact normal direction $\mathbf{n}_t$ is used to shape the covariance of the proposal.
The resulting proposal is given by
\begin{equation}
\pi(
\mathbf{U}_t
\mid
\bar{\mathbf{U}}_t,\mathcal{C}_t
)
=
\mathcal{N}
(
\bar{\mathbf{U}}_t+\boldsymbol{\mu}_t,
\boldsymbol{\Sigma}_t
).
\end{equation}
The following paragraphs formulate how $\boldsymbol{\mu}_t, \Sigma_t$ are computed.

\noindent\textbf{Contact-Normal Covariance Shaping:}
Contact reduces the utility of samples that continue to move along the active contact-normal direction. We therefore attenuate sampling variance along $\mathbf{n}_t$ while preserving tangential exploration. The confidence-dependent normal scale is
\begin{equation}
\gamma_t
=
1-c_t(1-\gamma_{\min}),
\end{equation}
where $\gamma_{\min}\in(0,1]$ determines the maximum covariance compression, which we set empirically to 0.1.
We use separate exploration scales for linear and angular velocity.
Linear velocity samples are drawn from a contact-normal-shaped anisotropic
Gaussian, while angular velocity samples are drawn from an isotropic Gaussian.

\label{sec:contact_representation}
\begin{equation}
\boldsymbol{\Sigma}_{v,t}
=
\sigma_v^2
\left[
\mathbf{I}_3
-
\left(1-\gamma_t^2\right)
\mathbf{n}_t\mathbf{n}_t^\top
\right],
\boldsymbol{\Sigma}_{\omega, t}
=
\sigma_\omega^2 \mathbf{I}_3
\end{equation}



\noindent\textbf{Search-Direction Mean Shift:}
Covariance shaping suppresses exploration along the contact axis but does not encode a signed preference for moving away from contact.
We therefore shift the proposal mean along the search direction,
\begin{equation}
\boldsymbol{\mu}_t
=
\alpha c_t\mathbf{s}_t,
\end{equation}
where $\alpha$ controls the magnitude of the search bias,
with the sampled control given by
\begin{equation}
\mathbf{u}_k^{(i)}
\sim
\mathcal{N}
\left(
\begin{bmatrix}
\bar{\mathbf{v}}_k + \alpha c_t \mathbf{s}_t \\
\bar{\boldsymbol{\omega}}_k
\end{bmatrix},
\begin{bmatrix}
\boldsymbol{\Sigma}_{v,t} & \mathbf{0} \\
\mathbf{0} & \boldsymbol{\Sigma}_{\omega, t}
\end{bmatrix}
\right).
\end{equation}


%

The nominal sequence is retained as the first rollout,
%
while the remaining candidates are sampled from the adaptive proposal and evaluated using the same task objective.  
\section{Experiments \& Results}
\begin{figure*}[t]
    \centering
    \vspace{3mm}
    \includegraphics[width=0.9\linewidth]{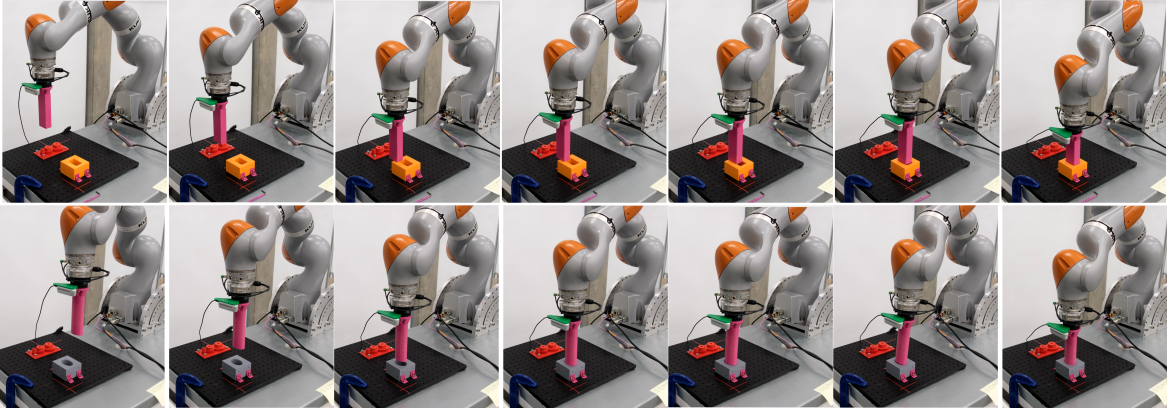}
    \caption{This figure shows qualitative results of the proposed adaptive kernel across \textsc{Rectangle, Ellipse} geometries. The sequences show the robot negotiating with the lip of the hole. Using contact information enables fine adaptation between the robot approach and insertion phases.}
    \label{fig:qualitative-results}
\end{figure*}

\begin{figure*}
    \centering
    \includegraphics[width=1.0\linewidth]{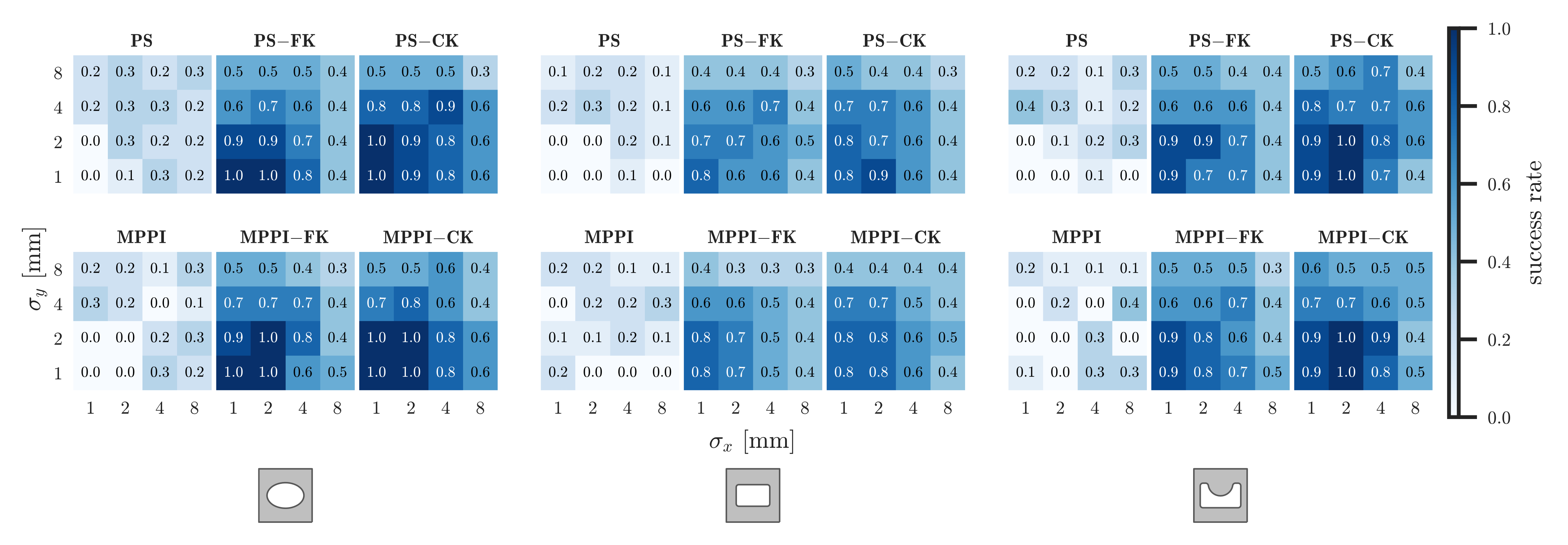}
    \caption{\textbf{Pose-error Sensitivity}: Sensitivity analysis for three variants of the two standard SMPC algorithms, \textbf{PS, MPPI} --- baseline isotropic Gaussian, anisotropic Gaussian dictated by the contact normal --- \textbf{FK}, proposed anisotropic Gaussian kernel dictated by the contact normal and escape direction --- \textbf{CK}. Each tile represents success ratios for 10 insertions attempts performed with random perturbations to the task goal. The perturbations sigmas $\sigma_x, \sigma_y$ are given along the horizontal and vertical axes respectively, with a perturbation along yaw fixed, $\sigma_\psi=0.04$. Higher success ratios indicate contact information enables both variants to adapt insertion strategy.}
    \label{fig:sensitivity-results}
\end{figure*}

We evaluate the proposed contact-adaptive proposal kernel on the contact-rich insertion task.
Experiments are setup to evaluate the following questions:
\begin{itemize}
    \item[\textbf{Q1}] Does contact-conditioned proposal adaptation improve insertion success over standard isotropic SMPC?
    \item[\textbf{Q2}] Does the contact-conditioned proposal adaptation improve time to insertion under the same sampling budget?
    \item[\textbf{Q3}] How does the insertion performance change as error in the tracked hole pose increases?
\end{itemize}
%
%

We evaluate these questions using controller simulation experiments and real-robot validation. 
First, we compare insertion success between the isotropic baseline, the contact-normal ablation (\textbf{FK}), and the ContactKernels (\textbf{CK}) under the same perturbation conditions to address \textbf{Q1}.
Second, we use the same trials to comapre time-to-isnertion under identical rollout budget tp address \textbf{Q2}.
Third, we stratify performance by controlled perturbations of the tracked hole pose to address \textbf{Q3}.
Finally, we repeat the comparisons on the real robot to test whether the improvements claimed to address \textbf{Q1} and \textbf{Q2} persist under sensing noise in real world settings.

\subsection{Experiment Setup}
All simulation experiments are evaluated via \textsc{MuJoCo}~\cite{todorov_mujoco_2012}.
For the real robot we evaluate the controller at 20Hz on a Kuka IIWA 7DoF arm, equipped with an ATI F/T sensor(100Hz).
An Operation Space Controller (OSC)~\cite{Khatib1987AFormulation} is used to translate computed end-effector twists to commanded robot state.
We report on insertion success and convergence time to evaluate individual methods.


We implement the baseline controllers \textbf{PS}~\cite{Howell2022PredictiveMuJoCo} and \textbf{MPPI}~\cite{theodorou_generalized_2010} with isotropic gaussian kernels using the \textsc{Judo} framework~\cite{li_judo_2025}.
Each baseline is compared with corresponding variants using the proposed adaptive kernels --- \textbf{PS-CK}, \textbf{MPPI-CK}.
We also evaluate a contact-normal ablation, denoted \textbf{FK}.
The comparison between the isotropic baseline and \textbf{FK} isolates the effect of contact-normal covariance shaping.
The comparison between \textbf{FK} and \textbf{CK} isolates the additional effect of contact-location estimation and the search direction mean shift.

The controllers are tested on 3 different geometries --- \textsc{Rectangle, Ellipse, Arch}.
These geometries expose the controller to different local contact structures, including the non-convex \textsc{Arch} geometry.
Insertions are evaluated on insertion success, which we define as $0.9d_{\text{goal}}$, in consonance with estimation error range of $0.1\text{mm}$.
We use the same rollout budget and controller parameters for all variants.
\cref{tab:controller-params} summarises various parameters used across the experiments.

\begin{table}[t]
\centering
\caption{Controller parameters used across experiments.}
\label{tab:controller-params}
\begin{tabular}{l|c}
\hline
\textbf{Parameter} & \textbf{Value} \\
\hline
Planning horizon ($H$) & $5.0$ s \\
Control frequency & $20$ Hz \\
Control knots ($M$) & $4$ \\
Rollouts ($K$) & $16$ \\
Linear Sampling Variance ($\sigma_v$) & $0.01$ \\
Angular Sampling Variance ($\sigma_{\omega}$) & $0.01$ \\
Normal covariance scale ($\gamma_{\min}$) & $0.1$ \\
Search bias ($\alpha$) & $0.005$ m/s \\
MPPI temperature ($\lambda$) & $0.05$ \\
Episode timeout & $75$ s \\
\hline
\end{tabular}
\end{table}


\subsection{Sensitivity Analysis}
To address \textbf{Q3}, we systematically perturb the hole pose supplied to the controller and evaluate insertion success and time-to-insertion in simulation.
We perturb the tracked target pose in $x, y, \psi$, while keeping the true simulated hole pose fixed.
For each perturbation cell, 10 insertions are performed with corresponding random seeds.
We aggregate insertion success (yes/no) and time-to-insertion (seconds) and report them in \cref{fig:sensitivity-summary-results}
We report insertion success results for each perturbation in \cref{fig:sensitivity-results}
This experiment measures how insertion success and time-to-insertion change as the tracked hole pose becomes less accurate.
Higher insertion success and lower time-to-insertion indicate superior performance.

\subsection{Real Robot Evaluation}
We use the real-robot experiment to test whether the trends in simulation persist under sensing noise and model mismatch.
On the real robot, we evaluate the MPPI and MPPI-CK variants.
A total of 10 insertions are attempted for each individual geometry and SMPC variant.
We report insertion success and time-to-insertion metrics for each trial in \cref{fig:real-robot-results}
\cref{tab:controller-params} reports controller parameter values that were used to run the real robot insertion experiments
As with the simulation comparisons, a higher insertion success rate and lower convergence time indicate superior performance.

\subsection{Results}

\begin{figure}
    \centering
    \includegraphics[width=1.0\linewidth]{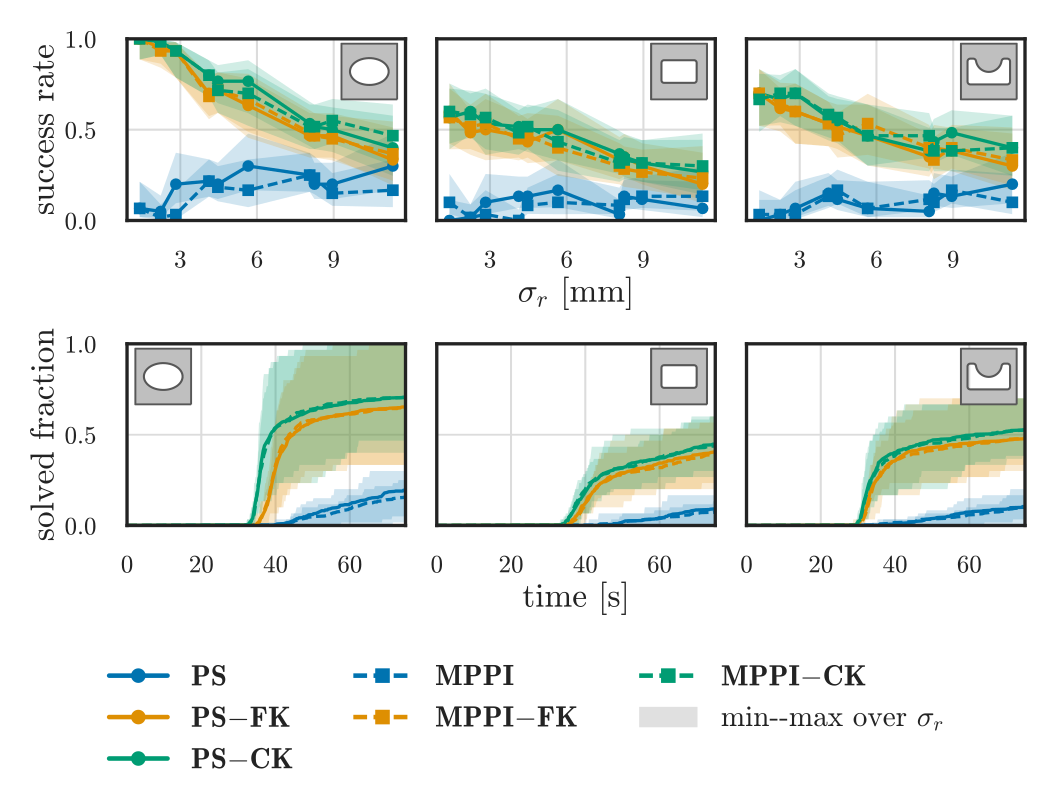}
    \caption{\textbf{Pose-error Sensitivity Summary}: This figure summarises epistemic sensitivity results the three specified geometries. \textbf{Top row}: Here $\sigma_r = \sqrt{\sigma_x^{2} + \sigma_y^{2}}$ specifies the noise radius, higher success rates indicate superior precision. \textbf{Bottom row}: Solved fraction indicates the fraction of insertion attempts that have succeeded vs. time in seconds. Higher solved fraction at any given time indicate insertion efficiency.}
    \label{fig:sensitivity-summary-results}
\end{figure}

\noindent\textbf{(Q1) Contact-conditioned sampling improves insertion success}:
sensitivity analysis in Figs.~\cref{fig:sensitivity-results} and~\cref{fig:sensitivity-summary-results} shows a substantial improvement from contact-conditioned sampling.
Averaged over the displayed perturbation conditions and three geometries, \textbf{PS-CK} achieves a success rate of 66.7\%, compared with 15.8\% for \textbf{PS}, corresponding to an improvement of 50.8 percentage points. 
Similarly, \textbf{MPPI-CK} achieves 65.6\%, compared with 13.5\% for \textbf{MPPI}, corresponding to an improvement of 52.1 percentage points.
The force-conditioned variants also improve substantially over the isotropic baselines, reaching 59.6\% for \textbf{PS-FK} and 59.0\% for \textbf{MPPI-FK}.
\textbf{CK} provides a further improvement of 7.1 percentage points over \textbf{FK} for \textbf{PS} and 6.7 percentage points for \textbf{MPPI}, indicating an additional benefit from incorporating the estimated contact location and search direction.

Insertion success improvement is consistent across three tested geometries and both SMPC control update laws.
For PS, \textbf{CK} increases the mean success rate from 20.6\% to 71.9\% on \textsc{ellipse}, from 11.3\% to 57.5\% on \textsc{rectangle}, and from 15.6\% to 70.6\% on \textsc{arch}.
For MPPI, the corresponding improvements are 15.0\% to 70.6\%, 12.5\% to 57.5\%, and 13.1\% to 68.8\%, respectively.
The similar gains observed for PS and MPPI suggest that the improvement arises from the proposed proposal kernel rather than from a particular SMPC control-update rule.

The real-robot experiments show the same trend.
MPPI-\textbf{CK} succeeds in 9/10, 10/10, and 8/10 trials on \textsc{rectangle}, \textsc{ellipse}, and \textsc{arch}, respectively, compared with 6/10, 5/10, and 5/10 for MPPI.
This corresponds to an increase in insertion success from 16/30 trials (53.3\%) to 27/30 trials (90.0\%).

\begin{figure}
    \centering
    \includegraphics[width=1.0\linewidth]{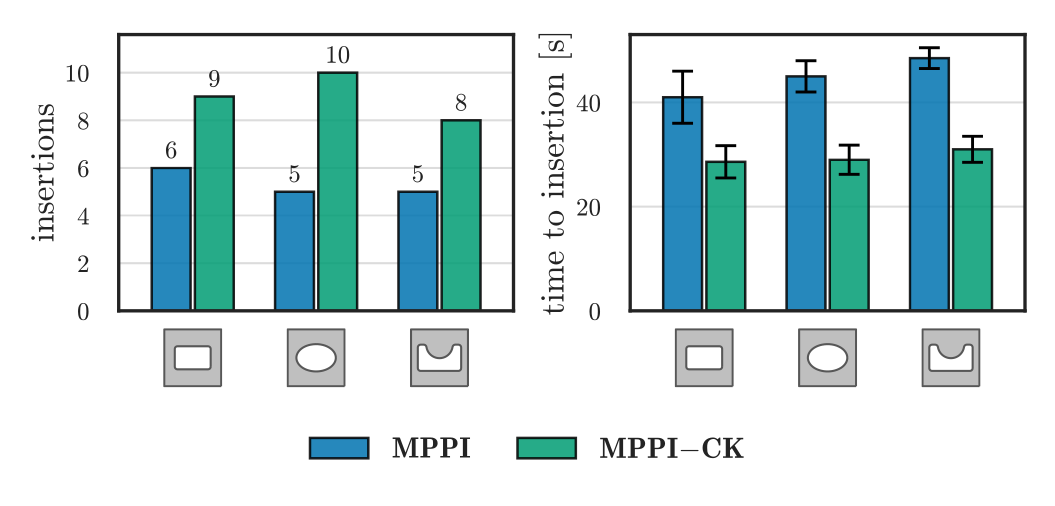}
    \caption{\textbf{Robot Experiments}: This figure summarises insertion and time-to-insertion results on the real robot setup.}
    \label{fig:real-robot-results}
\end{figure}

\noindent\textbf{(Q2) Contact-conditioned sampling improves time-to-insertion}:
The solved-fraction curves in \cref{fig:sensitivity-summary-results} show that the contact-conditioned variants reach successful insertion earlier than the isotropic baselines in simulation.
This trend is consistent across all three geometries and for both PS and MPPI.
FK improves convergence over the baseline, while CK achieves a similar or higher solved fraction throughout the episode.
MPPI-\textbf{CK} also reduces the mean time-to-insertion across all three geometries by $33\%$, as shown in~\cref{fig:real-robot-results}.

\noindent\textbf{(Q3) Contact conditioning reduces sensitivity to tracked hole-pose error}: \cref{fig:sensitivity-results} shows that the isotropic \textbf{PSA} and \textbf{MPPI} baselines achieve low success across most of the tested pose perturbations.
In contrast, \textbf{FK, CK} maintains substantially higher success over a wider range of translational and rotational errors.
The aggregated results in \cref{fig:sensitivity-summary-results} show that success decreases as the translational error magnitude increases, but the contact-conditioned variants retain higher success across the tested error range.
This shows that contact-conditioned sampling reduces the controller's dependence on an accurate tracked hole pose.

\section{Discussion}

Contact-conditioned sampling improves insertion by reallocating the finite rollout budget after unintended contact.
\textbf{FK} suppresses exploration along the contact normal, while \textbf{CK} also biases samples toward a corrective search direction.
This makes the sampled trajectories more consistent with the observed contact state. 
The similar gains for \textbf{PS} and \textbf{MPPI} suggest that the benefit comes from proposal adaptation rather than a specific control-update rule.

Contact is particularly useful when the tracked hole pose is inaccurate.
The nominal goal provides global task information but does not explain the local constraint created by unexpected contact.
The wrench-derived contact estimate supplies this local information at execution time and redirects exploration without changing the task objective or retraining a policy.

The current kernel is limited by its local contact representation.
The search direction works well for the convex geometries tested here, but may be insufficient for non-convex geometry such as \textsc{ARCH}.
The estimator also assumes a single dominant contact, so simultaneous contacts can make the wrench estimate ambiguous.
A natural extension is to combine local contact estimates with global geometry, such as a signed-distance field, to guide proposal adaptation.

\section{Conclusion}
We introduced \emph{ContactKernels}, a contact-conditioned proposal kernel that uses estimated contact geometry to guide exploration in sampling-based MPC.
On the real robot, MPPI-CK achieved 90.0\% overall insertion success, compared with 53.3\% for baseline MPPI, while also reducing mean time to insertion.
The consistent gains across Predictive Sampling and MPPI in simulation further suggest that these improvements arise from contact-conditioned exploration rather than a specific control update rule.
Overall, the results show that explicitly closing the loop between contact sensing and control exploration can improve the robustness and efficiency of contact-rich insertion.


\printbibliography

\end{document}